\documentclass[letterpaper]{article} 
\usepackage{aaai25}  
\usepackage{times}  
\usepackage{helvet}  
\usepackage{courier}  
\usepackage[hyphens]{url}  
\usepackage{graphicx} 
\usepackage{natbib}  
\usepackage{caption} 
\usepackage{algorithm}
\usepackage{algorithmic}

\usepackage{algorithm}
\usepackage{algorithmic}

\usepackage{framed,multirow}
\usepackage{array}
\usepackage{xcolor}
\usepackage{colortbl}
\usepackage{makecell}
\usepackage{amsmath}
\usepackage{amssymb}
\usepackage{dsfont}
\usepackage{threeparttable}
\usepackage{bbding}
\usepackage{pifont}
\DeclareMathOperator*{\argmax}{arg\,max}
\DeclareMathOperator*{\argmin}{arg\,min}
\newcommand{\xmark}{\text{\ding{55}} }

\usepackage{newfloat}
\usepackage{listings}
\DeclareCaptionStyle{ruled}{labelfont=normalfont,labelsep=colon,strut=off} 
\floatstyle{ruled}
\newfloat{listing}{tb}{lst}{}
\floatname{listing}{Listing}

\title{AIF-SFDA: Autonomous Information Filter-driven Source-Free Domain Adaptation for Medical Image Segmentation}
\author{
    Haojin Li\textsuperscript{\rm 1, \rm 2}, 
    Heng Li\textsuperscript{\rm 1, \footnote{Corresponding authors: lih3, liuj@sustech.edu.cn}}, 
    Jianyu Chen\textsuperscript{\rm 1, \rm 2}, 
    Rihan Zhong\textsuperscript{\rm 1, \rm 2}, 
    Ke Niu\textsuperscript{\rm 3}, 
    Huazhu Fu\textsuperscript{\rm 4}, 
    Jiang Liu\textsuperscript{\rm 1, \rm 2, \footnotemark[1]}
}
\affiliations{
    \textsuperscript{\rm 1} Research Institute of Trustworthy Adaptive Systems, Southern University of Science and Technology\\
    \textsuperscript{\rm 2} Department of Computer Science and Engineering, Southern University of Science and Technology\\
    \textsuperscript{\rm 3} Beijing Information Science \& Technology University\\
    \textsuperscript{\rm 4} Institute of High Performance Computing, Agency for Science, Technology and Research\\
}

\usepackage{bibentry}

\begin{document}

\maketitle

\renewcommand{\thefootnote}{}
\footnotetext{This work has been accepted to AAAI 2025. The final version will be published in the AAAI 2025 Proceedings.}
\renewcommand{\thefootnote}{\arabic{footnote}}

\begin{abstract}

Decoupling domain-variant information (DVI) from domain-invariant information (DII) serves as a prominent strategy for mitigating domain shifts in the practical implementation of deep learning algorithms. 
However, in medical settings, concerns surrounding data collection and privacy often restrict access to both training and test data, hindering the empirical decoupling of information by existing methods.
To tackle this issue, we propose an Autonomous Information Filter-driven Source-free Domain Adaptation (AIF-SFDA) algorithm, which leverages a frequency-based learnable information filter to autonomously decouple DVI and DII. 
Information Bottleneck (IB) and Self-supervision (SS) are incorporated to optimize the learnable frequency filter. 
The IB governs the information flow within the filter to diminish redundant DVI, while SS preserves DII in alignment with the specific task and image modality. 
Thus, the autonomous information filter can overcome domain shifts relying solely on target data.
A series of experiments covering various medical image modalities and segmentation tasks were conducted to demonstrate the benefits of AIF-SFDA through comparisons with leading algorithms and ablation studies.
The code is available at https://github.com/JingHuaMan/AIF-SFDA.

\end{abstract}

\section{Introduction}

In recent years, there has been considerable advancement in the field of medical image segmentation methods based on deep learning (DL)~\cite{li2024fd}. However, real-world scenarios frequently involve open datasets where the test data (target domain) is unseen and likely exhibits domain shifts in comparison to the training data (source domain)~\cite{guan2021domain}, attributed to practical variations in acquisition devices, patient demographics, image quality, and other variables. These domain shifts can significantly affect the performance of segmentation models on target domains~\cite{li2023frequency}. Therefore, effectively transferring segmentation models from the source domain to the target domains is both logical and essential to enhance the utilization of DL algorithms.

\begin{figure}[t]
\centering
\includegraphics[width=1\columnwidth]{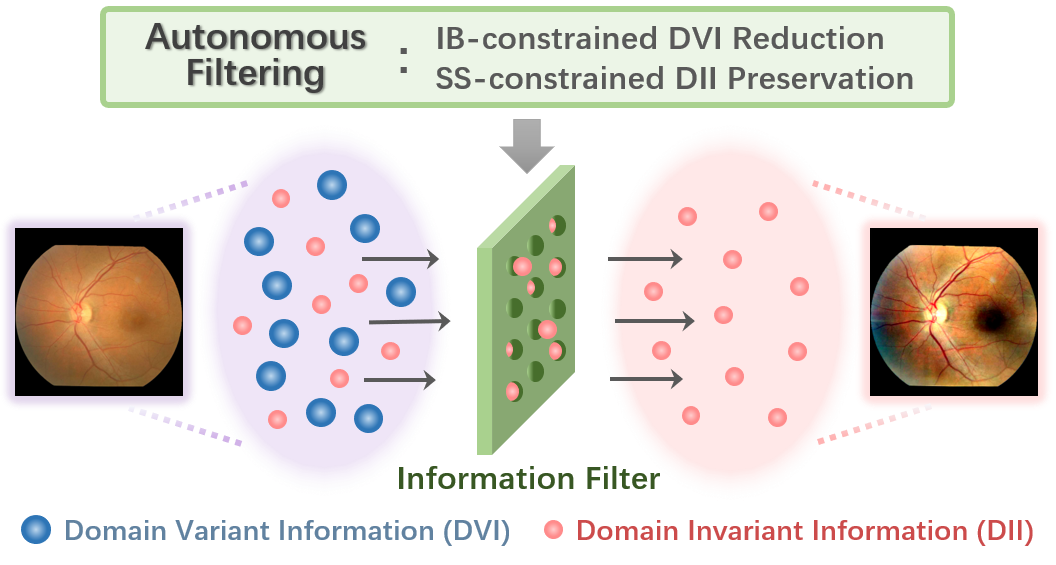}
\caption{
To develop an autonomous information filter in SFDA scenarios, we utilize IB-constrained mutual information constraint to reduce DVI in the image information while preserving DII through SS-constrained guidance.}
\label{fig:introduction}
\end{figure}

To address this issue, Unsupervised Domain Adaptation (UDA) has been proposed to generalize models by leveraging labeled source data in conjunction with unlabeled target data. A prominent strategy~\cite{yang2020fda,liu2021feddg} within UDA revolves around decoupling information into domain-variant and domain-invariant information (DVI \& DII), followed by the compression of DVI and the enhancement of DII to enable robust inference on novel data.

Specifically, configurable information filters based on frequency filtering are applied to process images or features, selecting DVI and DII from various frequency components.

In these type of approach, filter configurations are typically derived by identifying characteristics and commonalities between the source and target domains in the frequency spectrum. Configurations can be obtained by empirically comparing frequency features \cite{liu2023reducing} or through autonomous optimization guided by task-related losses \cite{lin2023deep}.
Nevertheless, jointly accessing both source and target domains leads to concerns involving data collection and privacy, often unacceptable in various practical contexts, especially medical scenarios.

Accordingly, source-free domain adaptation (SFDA)~\cite{li2024comprehensive} becomes imperative to enable the adaptation of pre-trained models solely using unlabeled target data.
However, challenges emerge in decoupling DVI and DII in SFDA settings. 
1) The absence of labeled source data in SFDA results in a lack of guidance for decoupling DVI and DII.
2) The preservation of DII is complicated when relying solely on unseen and unlabeled target data.
3) Frequency filter-driven algorithms often rely on empirical filtering configuration, which is impractical in SFDA.

To facilitate SFDA in medical image segmentation, we propose an Autonomous Information Filter (AIF-SFDA) autonomously aimed at decoupling DVI and DII for adaptation during the inference phase.
The AIF-SFDA enables learnable frequency filters through IB and SS, autonomously reducing DVI and preserving DII relying solely on target data.
Specifically, the IB regulates information flow within the filter to eliminate redundant DVI, while SS is derived from confidence-aware pseudo-labeling and consistency constraints to guide DII extraction.
Our main contributions can be summarized as follows:
\begin{itemize}
    \item We propose a source-free domain adaptation algorithm for medical image segmentation termed AIF-SFDA to adaptively decouple DVI and DII using learnable frequency filters.
    \item An autonomous information filter is constructed based on learnable frequency filters to reduce DVI and preserve DII, exclusively leveraging target data.
    \item The IB and SS are implemented to regulate the learnable frequency filters, enabling the adaptive decoupling of domain information for SFDA.
    \item Cross-domain experiments were conducted across diverse medical image modalities and segmentation tasks to assess the efficacy of AIF-SFDA through comparisons with state-of-the-art algorithms and ablation studies.
\end{itemize}

\section{Related Work}

\subsection{Source-free Domain Adaptation}

UDA algorithms have been designed to tackle domain shifts effectively by utilizing both the source and target data concurrently. However, due to concerns related to data access and privacy, SFDA has emerged as an alternative approach. SFDA allows the transfer of knowledge from a pre-trained source model to unlabeled target data without the need to access the source domain~\cite{li2024comprehensive}. SFDA methods can be broadly categorized into data-based and model-based approaches. Data-based methods process target data using knowledge from the source model to reduce discrepancies with the source domain, such as selecting target data to generate surrogate source data \cite{ye2021source} or using image translation to adapt target domain images to a source-like style \cite{yang2022source}. Model-based methods primarily involve self-supervised tasks on intermediate features and segmentation outputs, such as contrastive learning \cite{zhang2024improving}, pseudo-label supervision \cite{li2023toward}, or regularization constraints like entropy minimization \cite{fleuret2021uncertainty}.

\subsection{Frequency-based Domain Information Decoupling}

Frequency domain methods established by operations such as DCT and DFT enable data transformation between the spatial and frequency domains. Past studies on transfer learning have demonstrated that different frequency components exhibit varying domain-related characteristics, thus facilitating domain information decoupling \cite{xu2021fourier,liu2023reducing}. A common approach involves empirically designing filters based on the a priori knowledge contained in labeled source data. For instance, splitting DII and DVI domains by high/low frequencies with a fixed threshold \cite{yang2020fda,liu2021feddg,li2022domain} or selecting the most suitable components for the downstream task through spectral analysis \cite{huang2021fsdr}. These assumptions are typically derived from the labeled source data. Another approach involves using adaptive filters with learnable parameters to guide DII extraction through supervised task-related loss \cite{lin2023deep}. However, these methods that explicitly include supervised learning are not directly applicable in the SFDA scenario. Therefore, it is necessary to design a domain decoupling mechanism that does not require access to source data.

\subsection{Information Bottleneck in Deep learning}

IB theory is an information-theoretic approach \cite{tishby2000information}, aimed at obtaining compact data representations by reducing task-irrelevant parts of the data. For a certain model, IB achieves this by minimizing the mutual information (MI) between the input and intermediate variables, while maximizing MI between the intermediate variables and the output. Recently, IB theory has been used to provide interpretable analyses for DL methods due to its clear mathematical framework. \cite{tishby2015deep} viewed information extraction in multi-layer networks as deriving the minimum sufficient statistic, while \cite{kawaguchi2023does} showed that IB can control the generalization error of DL methods. Studies like \cite{alemi2016deep} explicitly utilize the IB principle, implementing a variational approximation with a variational network and showing high generalization performance. In this work, IB is applied to the information filter process to reduce DVI in the filtered image through MI constraints, thereby aiding the extraction of DII.

\begin{figure*}[t]
\centering
\includegraphics[width=2.1\columnwidth]{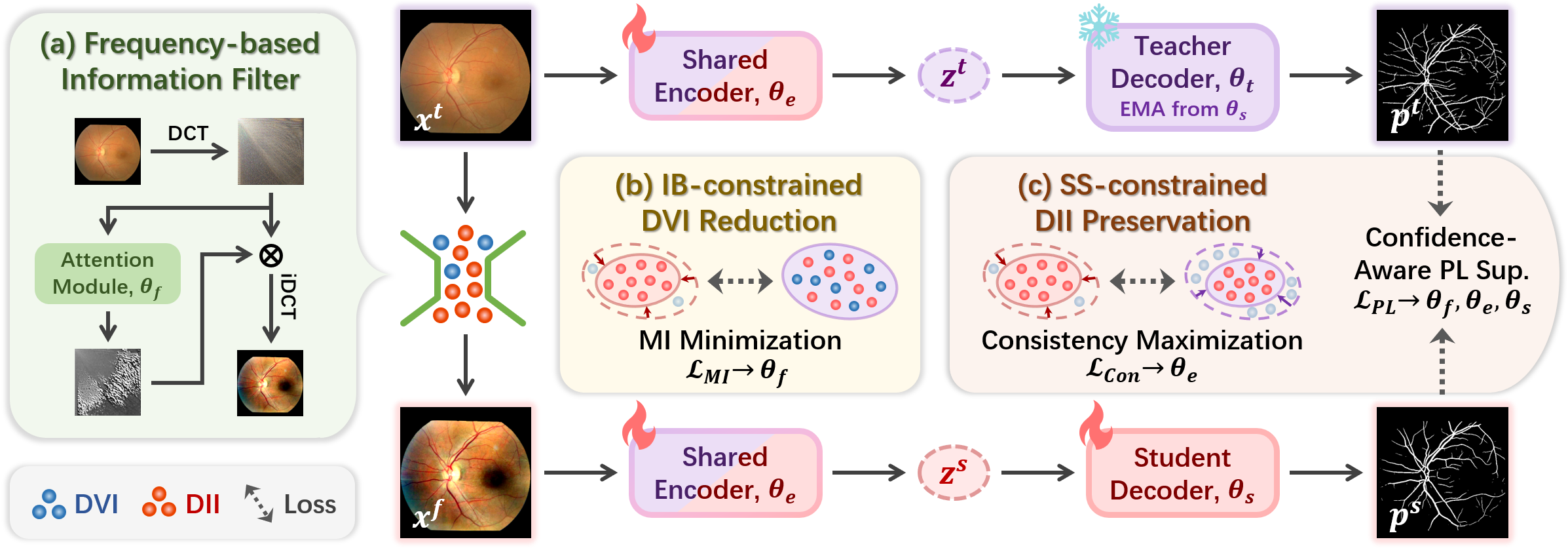}
\caption{The architecture of our proposed AIF-SFDA.}
\label{fig:architecture}
\end{figure*}

\section{Method}

\subsection{Overview}

To enhance the cross-domain performance of the segmentation model, SFDA consists of two distinct stages. In the source domain pre-training stage, given a source dataset $X_S=\{(x_i^s, y_i^s), i=1, \dots, N_S\}$ that includes images and corresponding labels, a source model $g(\cdot)$ is well-trained to obtain the parameter $\theta_o$, i.e., $\theta_o = \argmin_{\theta_o} \frac{1}{N_S} \sum_{i=1}^{N^s} l_s(g(x^s_i; \theta_o), y^s_i)$, where $l_s$ denotes a certain supervised segmentation loss. In the target domain adaptation stage, given an image-only unlabelled target dataset $X_T=\{(x^t_i,), i=1, \dots, N_T\}$ as well as the source model, it is necessary to improve the segmentation model's generalizability on the target data in the absence of direct access to source data.

The overall architecture of the AIF-SFDA algorithm we designed is shown in Figure \ref{fig:architecture}. To effectively boost the domain information decoupling through image transformation, we incorporate a frequency-based information filter $f(\cdot)$ that autonomously decouples DVI and DII in the target domain image $x^t$ according to task type and instance characteristics. To guarantee that the information filter works robustly on the target data, IB-constrained DVI reduction is firstly achieved by optimizing mutual information minimization loss $\mathcal{L}_{MI}$
based on the image features $z^s$ and $z^t$ before and after filtering. Furthermore, SS-constrained DII is preserved through confidence-aware pseudo-label supervision loss $\mathcal{L}_{PL}$ and adversarial feature consistency loss $\mathcal{L}_{Con}$, ultimately enhancing the generalization performance of the segmentation model.

\subsection{Frequency-based Information Filter}

To adaptively compare and select domain variant and invariant information, an adaptive filter should possess two properties: 1) learnable parameters that can be easily optimized, and 2) the ability to adaptively process image information based on the input image. To achieve these two goals, our proposed adaptive filtering can be briefly described as applying spatial attention to the spectrum obtained after the DCT transformation. Denote the 2D DCT process as $\mathcal{F}(\cdot)$, and the basis functions are:
\begin{equation}
B_{u,v}^{i,j}=\cos(\dfrac{\pi(2i+1)u}{2H})\cos(\dfrac{\pi(2j+1)v}{2W}).
\label{eq:dct_basis}
\end{equation}

For each channel of $x^t$, perform the 2D DCT process (assuming $x^t$ is a grayscale image):

\begin{equation}
\mathcal{F}(x^t)_{u,v}=\sum_{i=0}^{H-1}\sum_{j=0}^{W-1}x^t_{i,j}B_{u,v}^{i,j},
\label{eq:dct}
\end{equation}

where $(u,v)$ are the indices on the spectrum, $\mathcal{F}(x^t) \in \mathbb{R}^{H,W}$. Then, input $\mathcal{F}(x^t)$ into the attention module $M_{\theta_f}(\cdot)$ to obtain the attention map. The adaptive filtering process is formulated as:
\begin{equation}
x^f=f_{\theta_f}(x^t)=\mathcal{F}^{-1}(M_{\theta_f}(\mathcal{F}(x))\odot\mathcal{F}(x)),
\label{eq:adaptive_filtering}
\end{equation}

where $\mathcal{F}^{-1}(\cdot)$ denotes the inverse DCT transform, and $\odot$ represents the Hadamard product. Through the above process, a filter capable of self-adjusting to select and remove information based on the input image is established.

\subsection{Domain Information Decoupling driven by Adaptive Filtering}

\subsubsection{IB-constrained DVI Reduction}

In the flow of image segmentation algorithms that incorporate an information filter, $x^f$ can be interpreted as an intermediate variable. Referring to the common practice of IB theory, we use information-theoretic methods to constrain the feature embeddings of $x^t$ and $x^f$ in order to modulate the adaptive filter. This ensures that $x^f$ approximates as closely as possible the task-relevant minimal sufficient statistics of $x^t$, thereby reducing the unwanted DVI. This process is described mathematically by IB as:
\begin{equation}
\mathop{\min}_{p(x^f|x^t)} [I(x^f;x^t)-\beta I(\hat{y};x^f)],
\label{eq:mi_goal}
\end{equation}

where $\beta$ is a Lagrange multiplier.

Considering the difficulty of quantifying the latter term in Eq. \ref{eq:mi_goal}, we constrain only the former term. Since the computational complexity is limited by the high dimension of the data if MI is computed directly among images, we constrain it indirectly by the MI between the corresponding feature embeddings of $x^t$ and $x^f$ outputted by a shared encoder with parameter $\theta_e$. In other words, we aim to reduce $I(z^s;z^t)$. According to the definition in \cite{cheng2020club}, $I(z^s; z^t)$ has the following upper bound when $p(z^t | z^s)$ is known:
\begin{align}
I(z^s; z^t) &\leq \mathbb{E}_{p(z^s, z^t)}\big[\log p(z^t \mid z^s)\big] \notag \\
            &\quad - \mathbb{E}_{p(z^s)p(z^t)}\big[\log p(z^t \mid z^s)\big].
\label{eq:club_p_known}
\end{align}



However, since $p(z^t|z^s)$ is intractable, we approximate it with a variational distribution $q_{\theta_q}(z^t|z^s)$. When the approximation is good, we can reduce $I(z^s;z^t)$ by optimizing the following loss function:
\begin{equation}
\mathcal{L}_{MI}= \dfrac{1}{N}\sum_{i=1}^{N}[\log{q_{\theta_q}(z^t_i|z^s_i)}-\dfrac{1}{N}\sum_{j=1}^{N}\log{q_{\theta_q}(z^t_j|z^s_i)}],
\label{eq:loss_mi}
\end{equation}

To ensure that $q(z^t|z^s;\theta_q)$ can approximate $p(z^t|z^s)$, we need to include the negative log-likelihood loss function:
\begin{equation}
\mathcal{L}_{Li}= -\dfrac{1}{N}\sum_{i=1}^{N}\log{q_{\theta_q}(z^t|z^s)},
\label{eq:loss_likeli}
\end{equation}

The proposed IB-constrained DVI Reduction prompts $x^f$ to remove the useless information in $x^t$, but it does not guarantee the retention of the DII in the original data. Therefore, we need an information preservation mechanism to avoid the loss of meaningful information.

\subsubsection{SS-constrained DII Preservation}

To preserve DII in domain information decoupling, AIF-SFDA employs self-supervised learning by fully utilizing unlabelled target data, using pseudo-label (PL) supervision and feature consistency constraints.

PL self-supervision helps the information filter learn the most task-specific DII, while also enabling the segmentation model to adapt to changes in the filtered image.
A teacher-student architecture is employed to generate PLs while retaining source domain knowledge, with the teacher decoder parameterized by $\theta_t$ and the student decoder by $\theta_s$. Denote the outputs of teacher and student decoders as $p^t$ and $p^s$, the PL is $\smash{\hat{y^t} = \argmax_c{p_c^t}}$, and the confidence associated with the pseudo-label is $\smash{\phi(\hat{y^t}) = \max_c{p_c^t}}$.

To address the interference of low-confidence pixels in PLs on the optimization of the information filter and segmentation models, we implement a PL filtering mechanism based on a confidence threshold. This approach aims to mitigate the impact of potentially incorrect pseudo-labels. The PL supervised loss function is defined as follows:
\begin{equation}
\mathcal{L}_{PL}=\dfrac{1}{HW}\sum_{h,w}^{H,W} \mathds{1}[\phi(\hat{y}^t_{h,w})>\tau]l_{ce}(\hat{y}^t_{h,w},p^s_{h,w}),
\label{eq:loss_pseudo}
\end{equation}

where $\mathds{1}(\cdot)$ denotes the indicator function, $l_{ce}(\cdot,\cdot)$ represents pixel-wise cross entropy, and $\tau$ is the confidence threshold as a hyperparameter.

The distance between feature embeddings $z^s$ and $z^t$ should be maximized for the information filter, while it should be minimized for the segmentation model, as both embeddings contain the same task-specific semantic information. This dual requirement motivates the design of feature consistency constraints for optimizing the segmentation model encoder, which enhances both the information extraction of the segmentation model and the adversarial optimization of the information filter.

We use the commonly used cosine similarity to implement the consistency constraint:
\begin{equation}
\mathcal{L}_{Con}=\dfrac{\langle z^s, z^t \rangle}{||z^s||||z^t||},
\label{eq:loss_const}
\end{equation}
where $\langle \cdot, \cdot \rangle$ and $||\cdot||$ denotes inner product and L2 norm respectively.

\begin{algorithm}[tb]
\caption{The training procedures of AIF-SFDA}
\label{alg:procedure}
\textbf{Input}: Target dataset $X_T$, source model $g_{\theta_o}$, information filter $f_{\theta_f}$, variational distribution $q_{\theta_q}$, max training iteration number $N$\\
\begin{algorithmic}[1] 
\STATE $\theta_e, \theta_t \leftarrow \theta_o$ \hfill$\triangleright$ Copy to teacher model
\STATE $\theta_e, \theta_s \leftarrow \theta_o$ \hfill$\triangleright$ Copy to student model
\FOR{iter $k=1$ to $N$}
\STATE $x^s \sim X_T$. \hfill $\triangleright$ Sample target data
\STATE $x^f\leftarrow f_{\theta_f}(x^t)$. \hfill$\triangleright$ Eq. \ref{eq:adaptive_filtering}
\STATE $z^t, p^t \leftarrow g_{\theta_e,\theta_t}(x^t)$.\hfill$\triangleright$ Teacher model process $x^t$
\STATE $z^s, p^s \leftarrow g_{\theta_e,\theta_s}(x^f)$.\hfill$\triangleright$ Student model process $x^f$
\STATE Compute $\mathcal{L}_{MI}$, $\mathcal{L}_{Li}$, $\mathcal{L}_{Con}$ based on $z^t$ and $z^s$.
\STATE Compute $\mathcal{L}_{PL}$ based on $p^t$ and $p^s$.
\STATE Update $\theta_f$ by $\mathcal{L}_{PL}$ and $\mathcal{L}_{MI}$. \hfill$\triangleright$ Eq. \ref{eq:goal_1}
\STATE Update $\theta_e, \theta_s, \theta_q$ by $\mathcal{L}_{PL}$, $\mathcal{L}_{Li}$ and $\mathcal{L}_{Con}$. \hfill$\triangleright$ Eq. \ref{eq:goal_2}
\STATE Update $\theta_t$ based on $\theta_s$ through EMA.  \hfill$\triangleright$ Eq. \ref{eq:ema}
\ENDFOR
\end{algorithmic}
\end{algorithm}

\subsection{Source-Free Domain Adaptation}

The complete process of the proposed AIF-SFDA is outlined in Algorithm~\ref{alg:procedure}. The optimization process is divided into two steps. First, the information filter is optimized using pseudo-label self-supervision combined with IB-constrained DVI reduction, enabling the filter to autonomously extract DII and eliminate DVI:
\begin{equation}
\min_{\theta_f}[\mathcal{L}_{PL}+\alpha_1 \mathcal{L}_{MI}].
\label{eq:goal_1}
\end{equation}

Secondly, we optimize the parameters of the variational distributions in the student model and the MI constraints to assist in the optimization of the information filter and improve model generalization by learning the DII in the filtered image, i.e.:
\begin{equation}
\min_{\theta_e,\theta_s,\theta_q}[\mathcal{L}_{PL}+\alpha_2 \mathcal{L}_{Li}+\alpha_3 \mathcal{L}_{Con}],
\label{eq:goal_2}
\end{equation}

where $\alpha_1$, $\alpha_2$, and $\alpha_3$ are balancing hyperparameters. After each optimization iteration, we optimize the teacher decoder using the exponential moving average (EMA) based on the student decoder parameter:
\begin{equation}
\theta_t \leftarrow \eta \theta_t + (1-\eta) \theta_s,
\label{eq:ema}
\end{equation}

where $\eta$ is a coefficient ranging between $[0,1]$.

\section{Experiments}

\subsection{Experimental Settings}

\begin{table}[b]
\centering
\begin{tabular}{c|c|c}
\hline
Task    & Dataset & Volume \\ \hline
\makecell[c]{Retinal\\Vessel}  &  \makecell[c]{DRIVE\textsuperscript{*}, AVRDB,\\CHASEDB1, DRHAGIS,\\LES-AV, STARE} & \makecell[c]{40, 100,\\28, 40,\\22, 20} \\ \hline
\makecell[c]{Joint\\Cartilage} &  \makecell[c]{A\textsuperscript{*}, B, C} & \makecell[c]{956, 982, 750}     \\ \hline
\end{tabular}
\begin{tablenotes}
 \scriptsize
 \item{*} DRIVE and A are used as source domains in the two tasks, respectively. 
\end{tablenotes}
\caption{Datasets and their volumes used in this work.}
\label{tab:datasets}
\end{table}

\begin{table*}[!t]
\centering
\begin{tabular}{c|c|cc|cc|cc|cc|cc}
\hline
\multirow{2}{*}{Algorithm} & \multirow{2}{*}{SF\textsuperscript{*}}    & \multicolumn{2}{c|}{AVRDB} & \multicolumn{2}{c|}{CHASEDB1} & \multicolumn{2}{c|}{DRHAGIS} & \multicolumn{2}{c|}{LESAV} & \multicolumn{2}{c}{STARE} \\ \cline{3-12} 
\multicolumn{1}{c|}{}           &&  DSC$\uparrow$  & IoU$\uparrow$  & DSC$\uparrow$  & IoU$\uparrow$  & DSC$\uparrow$  & IoU$\uparrow$  &  DSC$\uparrow$  & IoU$\uparrow$  &  DSC$\uparrow$  & IoU$\uparrow$ \\ \hline
Source          &      /     &  54.34  &  39.47  &  52.36  &  37.87  &  54.65  &  39.48  &  58.12  &  42.18  &  58.53  &  42.69  \\ \hline
Rolling-Unet    &      /     &  59.64  &  43.23  &  59.95  &  43.08  &  61.91  &  44.91  &  63.44  &  46.51  &  64.28  &  47.52  \\
DTMFormer       &      /     &  60.73  &  44.54  &  60.69  &  44.33  &  62.22  &  45.76  &  63.84  &  47.45  &  64.64  &  48.41  \\
CS-CADA         &   \xmark   &  65.16  &  48.58  &  64.15  &  47.48  &  69.94  &  \textbf{55.56}  &  75.22  &  60.48  &  74.71  &  60.26  \\
DAMAN           &   \xmark   &  62.97  &  46.96  &  61.62  &  44.54  &  63.55  &  48.06  &  74.24  &  59.35  &  68.28  &  55.09  \\
MAAL            &   \xmark   &  60.80  &  47.73  &  61.80  &  45.72  &  68.88  &  52.60  &  77.56  &  \textbf{63.73}  &  70.95  &  58.97  \\
SFODA           & \checkmark &  64.69  &  48.79  &  63.88  &  46.87  &  66.25  &  51.03  &  76.83  &  62.69  &  74.69  &  61.65  \\ 
UPL-SFDA        & \checkmark &  61.31  &  43.57  &  62.88  &  45.81  &  66.64  &  51.49  &  76.59  &  62.45  &  75.33  &  62.20  \\
UBNA            & \checkmark &  61.56  &  43.83  &  63.35  &  46.28  &  66.37  &  51.17  &  76.86  &  62.71  &  74.61  &  61.54  \\
TSFCT           & \checkmark &  47.36  &  31.72  &  53.28  &  36.45  &  68.44  &  53.52  &  75.67  &  61.01  &  70.19  &  56.00  \\ \hline
AIF-SFDA        & \checkmark &  \textbf{66.22}  &  \textbf{49.85}  &  \textbf{64.44}  &  \textbf{47.49}  &  \textbf{69.99}  &  55.16  &  \textbf{78.08}  &  63.15  &  \textbf{76.68}  &  \textbf{63.01}  \\ \hline
\end{tabular}
\begin{tablenotes}
 \scriptsize
 \item{*} Here we denote vanilla, UDA, and SFDA segmentation algorithms by /, \xmark, and \checkmark, respectively. 
\end{tablenotes}
\caption{Comparison results on retinal vessel segmentation datasets, DSC (\%) and IoU (\%).}
\label{tab:comparison_vessel}
\end{table*}

\subsubsection{Datasets and Metrics}

The datasets involved in this work are shown in Table \ref{tab:datasets}. In the fundus photography retinal vessel segmentation task, all datasets used are publicly available, with DRIVE \cite{staal2004ridge} serving as the source domain and AVRDB \cite{niemeijer2011automatic}, CHASEDB1 \cite{owen2009measuring}, DRHAGIS \cite{holm2017dr}, LES-AV \cite{fraz2014ensemble}, STARE \cite{hoover2000locating} as the target domains. For the ultrasound joint cartilage segmentation task, the private datasets A, B, and C were provided by Southern University of Science and Technology Hospital, with dataset A used as the source domain. All datasets were randomly divided into training and test sets in a 1:1 ratio. We employed two common segmentation metrics to evaluate the performance of each algorithm: the Dice Similarity Coefficient (DSC) and the Intersection-over-Union (IoU), where higher DSC and IoU indicate better segmentation results.

\subsubsection{Implementation Details}

In the source domain pre-training stage, the segmentation model selects parameters at the optimal epoch based on the test set according to the early stopping mechanism. In the target domain adaptation stage, the model uses the parameters from the last epoch for performance evaluation on the test set. During the training process of both stages, we employ the Adam optimizer with an initial learning rate of 0.001 and a batch size of 2. The model is trained for 400 epochs in the retinal vessel segmentation task and 20 epochs in the ultrasound cartilage segmentation task, with the learning rate uniformly reduced to 0 in the latter half of the epochs. We use a naive U-net \cite{ronneberger2015u} as the segmentor, and the attention module $M$ in the information filter is implemented using a three-layer lightweight U-net. The variational distribution model $q$ employs a multivariate Gaussian distribution, parameterized using two 2-layer multi-layer perceptrons, each with a hidden size of 1024. The pseudo-label screening threshold $\tau$ is set to 0.8. The balancing coefficients $\alpha_1$, $\alpha_2$, and $\alpha_3$ are set to 0.5, 1, and 1, respectively. The EMA coefficient $\eta$ is set to 0.9995.

\begin{figure*}[t]
\centering
\includegraphics[width=2.1\columnwidth]{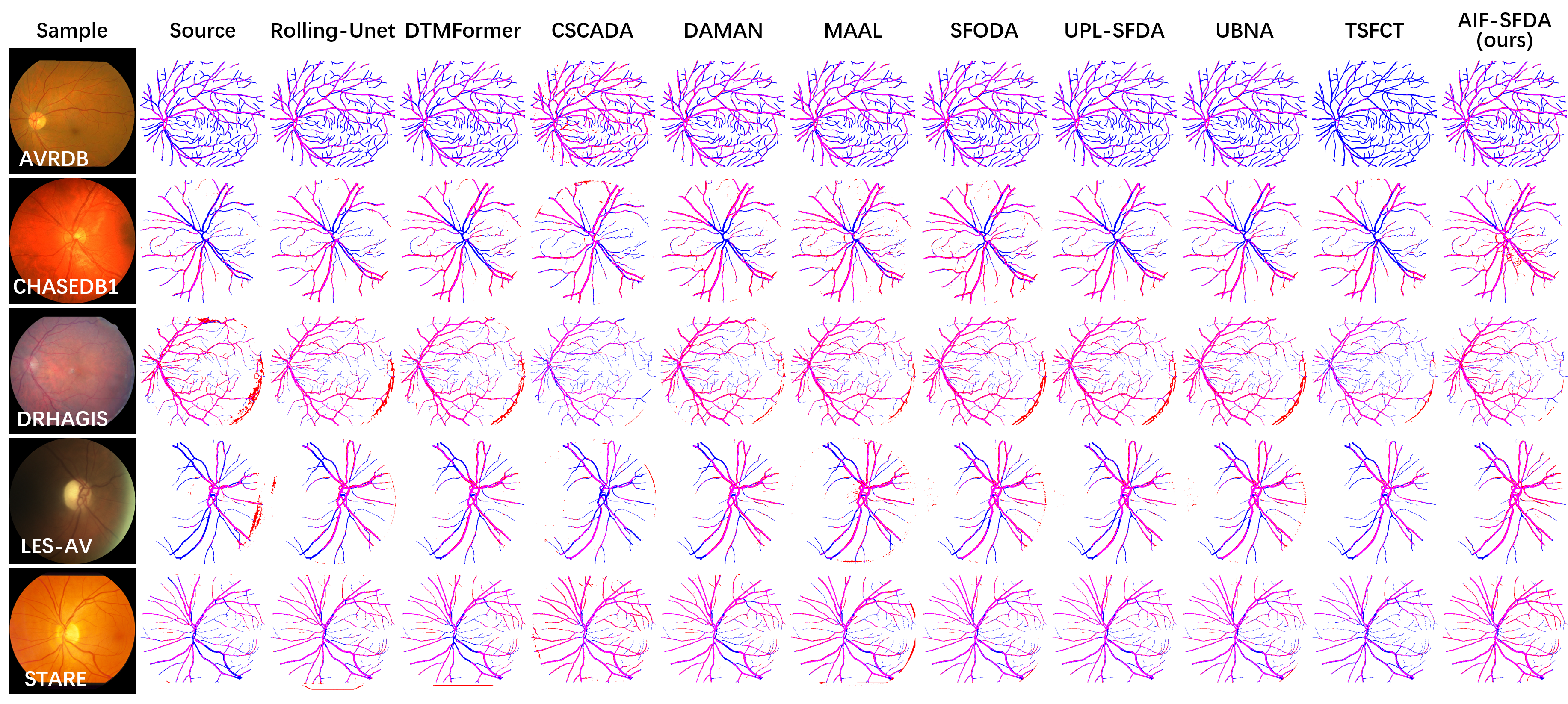}
\caption{Qualitative results for retinal vessel segmentation, where true positive pixels are colored in magenta, false positive pixels in red, and false negative pixels in blue.}
\label{fig:comparison_vessel}
\end{figure*}

\subsection{Comparison with State-of-the-Art Methods}

In the comparison experiments, we selected nine SOTA segmentation baselines, including two vanilla segmentation algorithms: Rolling-Unet \cite{liu2024rolling} and DTMFormer \cite{wang2024dtmformer}, three UDA algorithms: DAMAN \cite{mukherjee2022domain}, CS-CADA \cite{gu2022contrastive} and MAAL \cite{zhou2023unsupervised}, and four SFDA algorithms: SFODA \cite{niloy2024source}, UPL-SFDA \cite{wu2023upl}, UBNA \cite{klingner2022unsupervised} and TSFCT \cite{li2023toward}. 

\subsubsection{Result for Retinal Vessel Segmentation}

Table \ref{tab:comparison_vessel} presents the comparative results for the fundus vessel segmentation task, encompassing both the source domain model and the cross-domain performance of each baseline model, alongside our proposed AIF-SFDA. As observed, most baselines exhibit superior generalization on the target domain relative to the source model that solely employs  U-net. In general, UDA and SFDA algorithms outperform naive segmentation methods due to their specialized design for cross-domain segmentation. Among the UDA algorithms, CS-CADA outperforms DAMAN and MAAL, indicating that its approach could be well-suited for the vessel segmentation task. While most SFDA methods enhance cross-domain segmentation performance, with SFODA achieving over a $6\%$ DSC improvement across all datasets compared to the source model, TSFCT also demonstrates negative adaptation, underscoring the inherent challenges of adapting without access to source data. Notably, AIF-SFDA outperforms all baselines in generalizability, thereby validating the efficacy of the proposed autonomous information filtering mechanism.

\begin{table}[!t]
\centering
\begin{tabular}{c|cc|cc}
\hline
\multirow{2}{*}{Algorithm}    & \multicolumn{2}{c|}{B} & \multicolumn{2}{c}{C} \\ \cline{2-5} 
                              &  DSC$\uparrow$  & IoU$\uparrow$  & DSC$\uparrow$  & IoU$\uparrow$ \\ \hline
Source      &  63.43  &  57.98  &  51.57  &  52.20  \\ \hline
SFODA       &  66.69  &  60.11  &  55.14  &  55.21  \\
UPL-SFDA    &  64.08  &  60.44  &  49.75  &  44.86  \\
UBNA        &  65.39  &  60.57  &  54.48  &  53.36  \\
TSFCT       &  67.03  &  57.13  &  51.02  &  47.27  \\ \hline
AIF-SFDA    &  \textbf{69.14}  &  \textbf{62.07}  &  \textbf{55.16}  &  \textbf{55.70}  \\ \hline
\end{tabular}
\caption{Comparison results on ultrasound cartilage datasets, DSC (\%) and IoU (\%)}
\label{tab:comparison_ultrasound}
\end{table}

We show a qualitative comparison of fundus vessel segmentation experiments in Figure \ref{fig:comparison_vessel}. It can be seen that the domain differences between the fundus image datasets are mainly in the overall brightness of the images and task-irrelevant noise, which makes the baseline methods prone to ignoring small vessels in regions with uneven brightness and to misclassifying non-vessel noise pixels as false positives. Our proposed AIF-SFDA effectively exploits the frequency domain properties shared by vessel pixels, making the foreground pixels more conspicuous by processing the image with the autonomous information filter, thus improving the accuracy of small vessel segmentation and reducing the interference caused by unseen image noise in the target domains.

\begin{figure}[t]
\centering
\includegraphics[width=1.0\columnwidth]{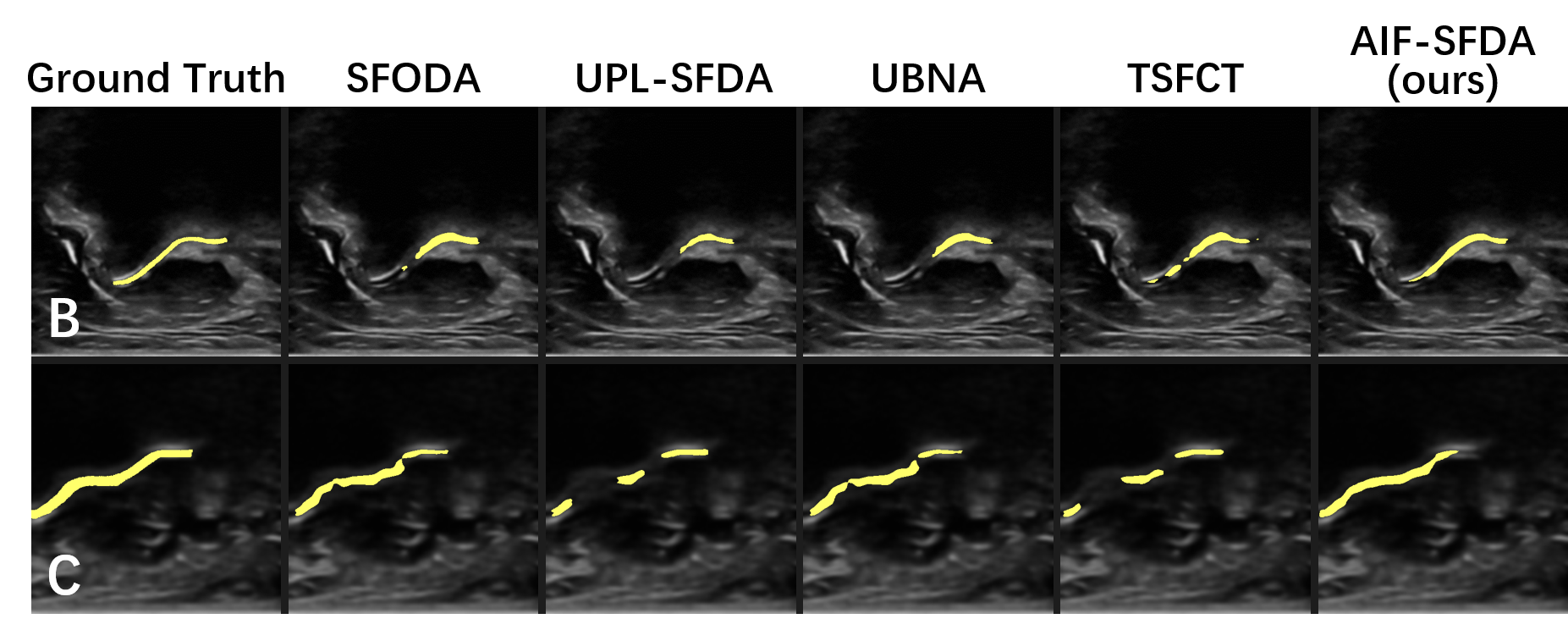}
\caption{Qualitative results for cartilage segmentation.}
\label{fig:comparison_ultrasound}
\end{figure}

\subsubsection{Result for Joint Cartilage Segmentation}

To conduct a more extensive comparison of medical images across multiple modalities, the segmentation results of the ultrasound joint cartilage segmentation task are shown in Table \ref{tab:comparison_ultrasound}. It can be found that the DSC of the source domain model on C is worse than the performance on dataset B, and both UPL-SFDA and TSFCT exhibit negative adaptation, possibly because C differs more from the source domain than B. Notably, TSFCT and SFODA achieve the best DSC in the baseline on B and C, respectively, suggesting that pseudo-labeled SFDA may perform better with smaller domain shifts, while feature-based algorithms are more effective with larger shifts. AIF-SFDA, which integrates both techniques, achieves the optimal DSC on both datasets.

Qualitative results are shown in Figure \ref{fig:comparison_ultrasound}. Generalization errors in cartilage segmentation often arise from contrast and luminance differences between ultrasound datasets, leading to poor continuity in segmentation. In dataset B, uneven image brightness hindered baseline methods from detecting end cartilage pixels, while the darker images in dataset C resulted in higher false negatives for SFDA baselines except for SFODA and UBNA. AIF-SFDA, using its information filter, mitigates the effects of blurring and low luminance, contributing to its strong generalization performance.

\begin{table}[!t]
\centering
\begin{tabular}{ccc|c|c}
\hline
 MI Min.      & PL Sel.      & Cons.       & DSC$\uparrow$  & IoU$\uparrow$  \\ \hline
              &              &              &  58.62  &  42.79  \\
 $\checkmark$ &              &              &  62.34  &  46.22  \\
 $\checkmark$ &              & $\checkmark$ &  65.25  &  49.08  \\
 $\checkmark$ & $\checkmark$ &              &  63.22  &  46.88  \\ \hline
 $\checkmark$ & $\checkmark$ & $\checkmark$ &  \textbf{66.22}  &  \textbf{49.85}  \\ \hline
\end{tabular}
\caption{Ablation Study on AVRDB.}
\label{tab:ablation_study}
\end{table}

\begin{figure}[t]
\centering
\includegraphics[width=0.9\columnwidth]{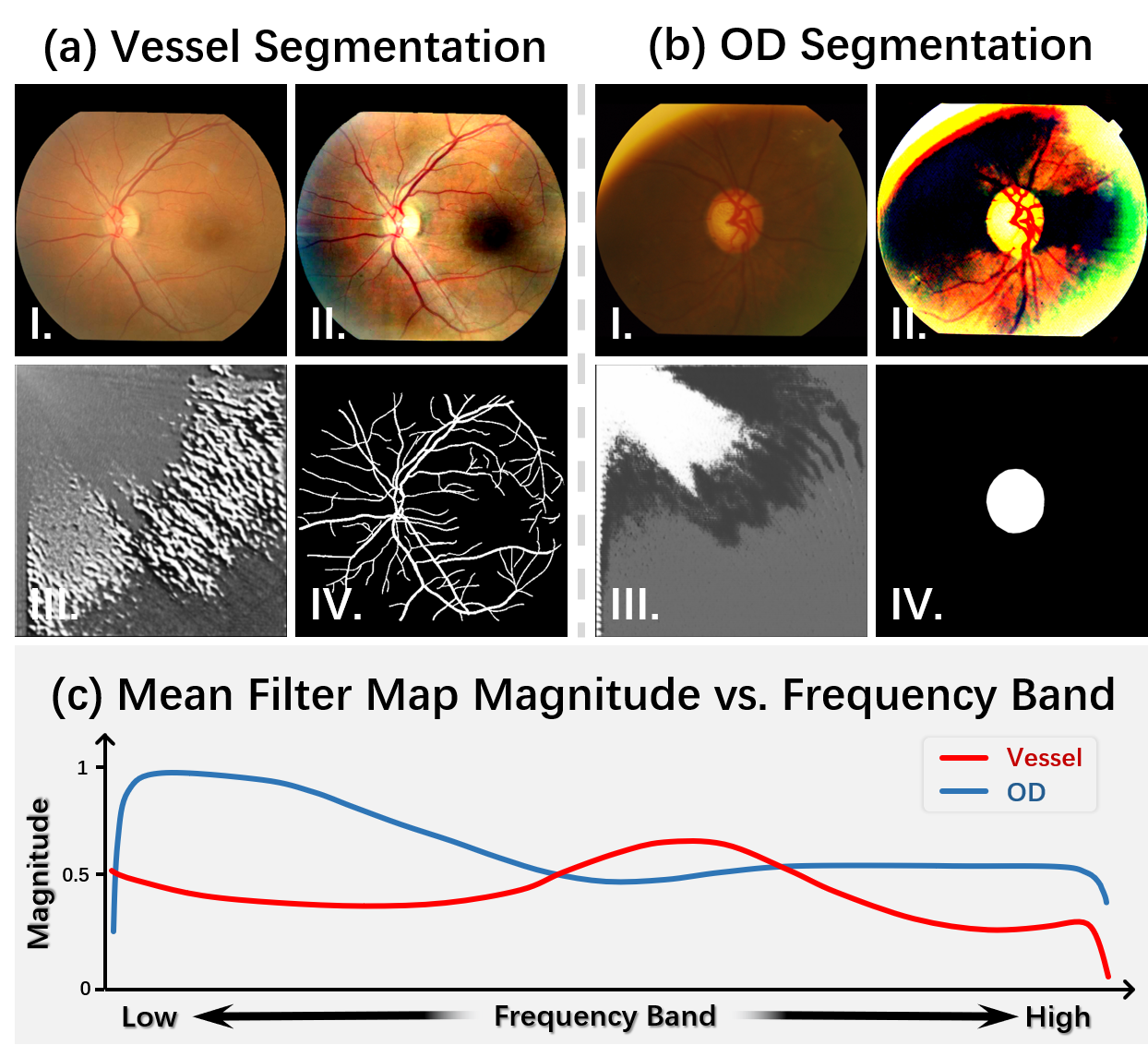}
\caption{
Information filter processing for different tasks. (a) and (b) show the filtering process for vessel and optic disc (OD) segmentation, respectively. (I. original image, II. filtered image, III. filter map, IV. ground truth)}
\label{fig:multitask}
\end{figure}

\subsection{Ablation Study}

\subsubsection{Ablation Study of Modules}

Table \ref{tab:ablation_study} shows the ablation experiments performed. We combine the MI minimization constraint (MI Min.) in IB-constrained DVI Reduction, the confidence-aware pseudo-label selection mechanism in SS-constrained DII Preservation (PL Sel.), and feature consistency constraints in SS-constrained DII Preservation (Cons.) sequentially to the model to validate the contribution of each module in AIF-SFDA to the target domain adaptation. The results indicate that the MI minimization constraint significantly contributes to enhancing the generalization of AIF-SFDA, confirming the effectiveness of selectively reducing DVI for information decoupling. Additionally, the incorporation of a confidence-aware pseudo-label selection mechanism markedly improves the stability of algorithm adaptation, while the feature consistency constraint enhances the model's feature extraction capability. The results of the ablation experiments prove the rationality of each module's settings.

\begin{figure}[t]
\centering
\includegraphics[width=1.\columnwidth]{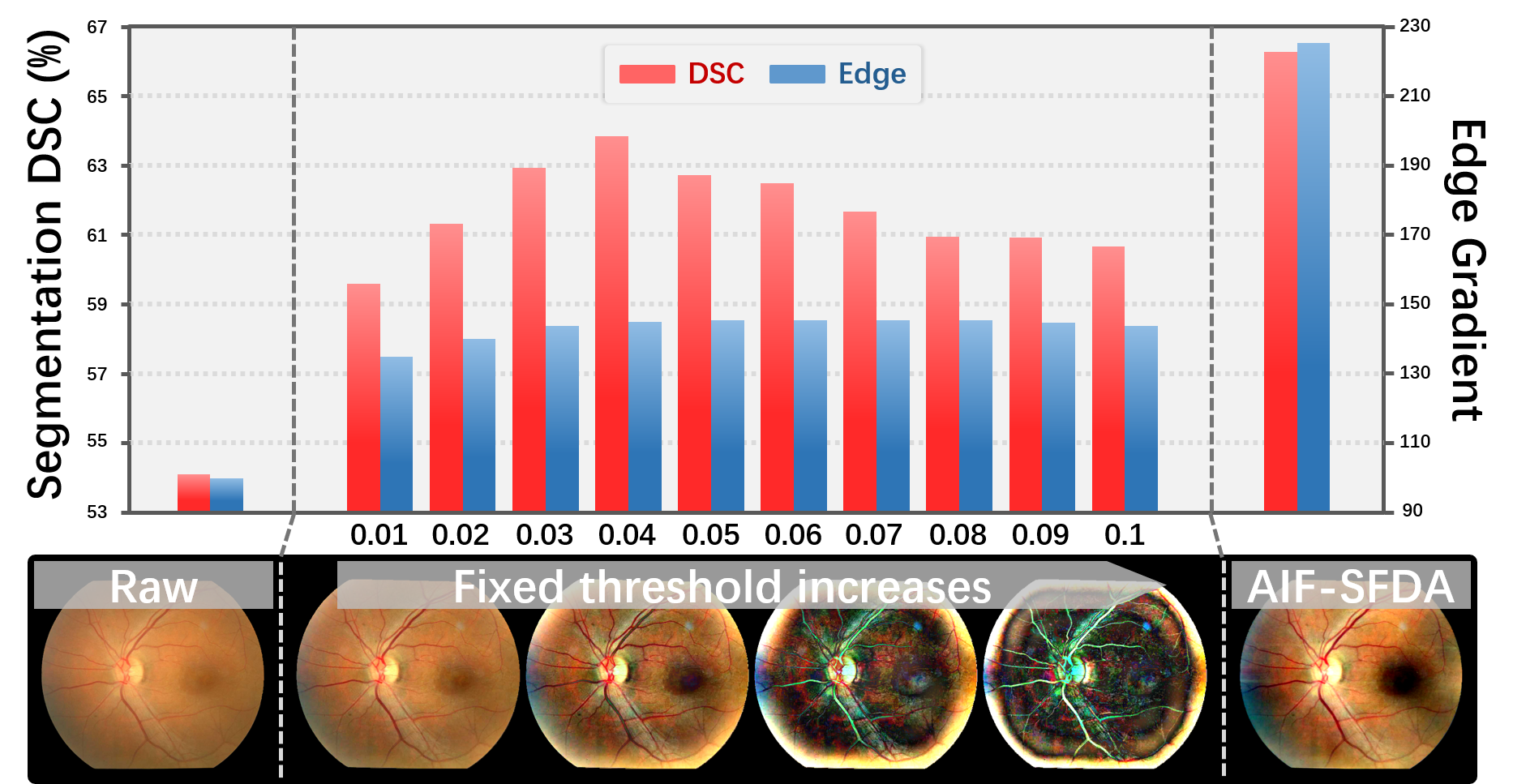}
\caption{
Comparison of fixed and adaptive filters on AVRDB. Left y-axis: DSC for vessel segmentation, indicating cross-domain performance. Right y-axis: Edge gradient magnitude around vessel pixels, indicating boundary distinctness. The filtered images outputted by fixed and adaptive filters are also visualized.}
\label{fig:fix_filter}
\end{figure}

\subsubsection{Information Filter in Various Tasks}

When faced with different segmentation tasks, we expect the information filter to extract the frequency domain components that best fit the DII for different segmentation objectives. To verify this, we included an extra fundus optic disc (OD) segmentation experiment (see Technical Appendix for experimental details and quantitative analysis). Figure \ref{fig:multitask} shows the performance differences of the information filter guided by different segmentation tasks, all within the modality of fundus photography. It is evident that the filter map focuses on the middle and high-frequency regions for vessels with more pronounced high-frequency characteristics, while for the OD segmentation task, the filter map focuses on the middle and low frequencies. This demonstrates the flexibility of the information filter in AIF-SFDA, which can self-adjust according to the task type and instance characteristics.

\subsubsection{Comparison with Fixed-setting Filters}

To demonstrate the importance of adaptive filtering for domain information decoupling, we replaced the information filter in AIF-SFDA with fixed frequency domain filters, removing the MI minimization loss for $\theta_f$ optimization. As shown in Figure \ref{fig:fix_filter}, following \cite{li2023frequency}, we used high-pass frequency filters with thresholds ranging from $0.01$ to $0.1$ for vessel segmentation on AVRDB. Fixed filter decoupling usually classifies components within the same frequency band as the same type of domain information and lacks the ability to autonomously adjust the filtering process based on the image, which prevents achieving optimal configuration. The learnable autonomous information filter in AIF-SFDA, compared to the fixed filter, increases vessel pixel edge gradients and effectively prevents artifact generation, enhancing cross-domain segmentation performance.

\section{Conclusion}

In this paper, we present an Autonomous Information Filter driven Source-free Domain Adaptation (AIF-SFDA) algorithm for medical image segmentation tasks. The method employs a frequency-based information filter to autonomously eliminate DVI from images through mutual information minimization based on information bottleneck theory and guides DII extraction through unsupervised task-relevant loss, thereby facilitating target domain adaptation. The results of cross-domain experiments on various medical image segmentation tasks demonstrate that AIF-SFDA outperforms existing SFDA methods.

\section{Acknowledgments}
This work was supported in part by the Basic Research Fund in Shenzhen Natural Science Foundation (Grant No. JCYJ20240813095112017), National Natural Science Foundation of China (Grant No. 62401246, 82272086), National Key R\&D Program of China (Grant No. 2024YFE0198100), and Shenzhen Medical Research Fund (Grant No.D2402014). 

\bibliography{aaai25}

\begin{thebibliography}{36}
\providecommand{\natexlab}[1]{#1}

\bibitem[{Alemi et~al.(2016)Alemi, Fischer, Dillon, and Murphy}]{alemi2016deep}
Alemi, A.~A.; Fischer, I.; Dillon, J.~V.; and Murphy, K. 2016.
\newblock Deep variational information bottleneck.
\newblock \emph{arXiv preprint arXiv:1612.00410}.

\bibitem[{Cheng et~al.(2020)Cheng, Hao, Dai, Liu, Gan, and Carin}]{cheng2020club}
Cheng, P.; Hao, W.; Dai, S.; Liu, J.; Gan, Z.; and Carin, L. 2020.
\newblock Club: A contrastive log-ratio upper bound of mutual information.
\newblock In \emph{International conference on machine learning}, 1779--1788. PMLR.

\bibitem[{Fleuret et~al.(2021)}]{fleuret2021uncertainty}
Fleuret, F.; et~al. 2021.
\newblock Uncertainty reduction for model adaptation in semantic segmentation.
\newblock In \emph{Proceedings of the IEEE/CVF Conference on Computer Vision and Pattern Recognition}, 9613--9623.

\bibitem[{Fraz et~al.(2014)Fraz, Barman, Remagnino, Hoppe, Rudnicka, and Owen}]{fraz2014ensemble}
Fraz, M.~M.; Barman, S.~A.; Remagnino, P.; Hoppe, A.; Rudnicka, A.~R.; and Owen, C.~G. 2014.
\newblock Ensemble classification-based vessel segmentation in retinal images using a novel vesselness enhancement method.
\newblock \emph{Annals of the British Machine Vision Association}, 2014: 32--32.

\bibitem[{Gu et~al.(2022)Gu, Zhang, Wang, Lei, Song, Zhang, Li, and Zhang}]{gu2022contrastive}
Gu, R.; Zhang, J.; Wang, G.; Lei, W.; Song, T.; Zhang, X.; Li, K.; and Zhang, S. 2022.
\newblock Contrastive semi-supervised learning for domain adaptive segmentation across similar anatomical structures.
\newblock \emph{IEEE Transactions on Medical Imaging}, 42(1): 245--256.

\bibitem[{Guan and Liu(2021)}]{guan2021domain}
Guan, H.; and Liu, M. 2021.
\newblock Domain adaptation for medical image analysis: a survey.
\newblock \emph{IEEE Transactions on Biomedical Engineering}, 69(3): 1173--1185.

\bibitem[{Holm et~al.(2017)Holm, Russell, Nourrit, and McLoughlin}]{holm2017dr}
Holm, S.; Russell, G.; Nourrit, V.; and McLoughlin, N. 2017.
\newblock DR HAGIS—a fundus image database for the automatic extraction of retinal surface vessels from diabetic patients.
\newblock \emph{Journal of Medical Imaging}, 4(1): 014503--014503.

\bibitem[{Hoover, Kouznetsova, and Goldbaum(2000)}]{hoover2000locating}
Hoover, A.; Kouznetsova, V.; and Goldbaum, M. 2000.
\newblock Locating blood vessels in retinal images by piecewise threshold probing of a matched filter response.
\newblock \emph{IEEE Transactions on Medical Imaging}, 19(3): 203--210.

\bibitem[{Huang et~al.(2021)Huang, Guan, Xiao, and Lu}]{huang2021fsdr}
Huang, J.; Guan, D.; Xiao, A.; and Lu, S. 2021.
\newblock Fsdr: Frequency space domain randomization for domain generalization.
\newblock In \emph{Proceedings of the IEEE/CVF conference on computer vision and pattern recognition}, 6891--6902.

\bibitem[{Kawaguchi et~al.(2023)Kawaguchi, Deng, Ji, and Huang}]{kawaguchi2023does}
Kawaguchi, K.; Deng, Z.; Ji, X.; and Huang, J. 2023.
\newblock How does information bottleneck help deep learning?
\newblock In \emph{International Conference on Machine Learning}, 16049--16096. PMLR.

\bibitem[{Klingner et~al.(2022)Klingner, Term{\"o}hlen, Ritterbach, and Fingscheidt}]{klingner2022unsupervised}
Klingner, M.; Term{\"o}hlen, J.-A.; Ritterbach, J.; and Fingscheidt, T. 2022.
\newblock Unsupervised batchnorm adaptation (ubna): A domain adaptation method for semantic segmentation without using source domain representations.
\newblock In \emph{Proceedings of the IEEE/CVF Winter Conference on Applications of Computer Vision}, 210--220.

\bibitem[{Li et~al.(2024{\natexlab{a}})Li, Li, Zhang, Li, Chen, Pan, Chen, and Liu}]{li2024fd}
Li, B.; Li, H.; Zhang, Y.; Li, H.; Chen, J.; Pan, F.; Chen, J.; and Liu, J. 2024{\natexlab{a}}.
\newblock FD-SDG: Frequency Dropout Based Single Source Domain Generalization Framework for Retinal Vessel Segmentation.
\newblock In \emph{International Conference on Intelligent Computing}, 393--404. Springer.

\bibitem[{Li et~al.(2022)Li, Li, Qiu, Hu, and Liu}]{li2022domain}
Li, H.; Li, H.; Qiu, Z.; Hu, Y.; and Liu, J. 2022.
\newblock Domain Adaptive Retinal Vessel Segmentation Guided by High-frequency Component.
\newblock In \emph{International Workshop on Ophthalmic Medical Image Analysis}, 115--124. Springer.

\bibitem[{Li et~al.(2023{\natexlab{a}})Li, Li, Zhao, Fu, Su, Hu, and Liu}]{li2023frequency}
Li, H.; Li, H.; Zhao, W.; Fu, H.; Su, X.; Hu, Y.; and Liu, J. 2023{\natexlab{a}}.
\newblock Frequency-mixed single-source domain generalization for medical image segmentation.
\newblock In \emph{International Conference on Medical Image Computing and Computer-Assisted Intervention}, 127--136. Springer.

\bibitem[{Li et~al.(2024{\natexlab{b}})Li, Yu, Du, Zhu, and Shen}]{li2024comprehensive}
Li, J.; Yu, Z.; Du, Z.; Zhu, L.; and Shen, H.~T. 2024{\natexlab{b}}.
\newblock A comprehensive survey on source-free domain adaptation.
\newblock \emph{IEEE Transactions on Pattern Analysis and Machine Intelligence}.

\bibitem[{Li et~al.(2023{\natexlab{b}})Li, Li, Luo, Zhou, Zhu, Xu, Yang, Wu, and Chen}]{li2023toward}
Li, Z.; Li, C.; Luo, X.; Zhou, Y.; Zhu, J.; Xu, C.; Yang, M.; Wu, Y.; and Chen, Y. 2023{\natexlab{b}}.
\newblock Toward source-free cross tissues histopathological cell segmentation via target-specific finetuning.
\newblock \emph{IEEE Transactions on Medical Imaging}, 42(9): 2666--2677.

\bibitem[{Lin et~al.(2023)Lin, Zhang, Huang, Lu, Lan, Chu, You, Wang, Liu, Parulkar et~al.}]{lin2023deep}
Lin, S.; Zhang, Z.; Huang, Z.; Lu, Y.; Lan, C.; Chu, P.; You, Q.; Wang, J.; Liu, Z.; Parulkar, A.; et~al. 2023.
\newblock Deep frequency filtering for domain generalization.
\newblock In \emph{Proceedings of the IEEE/CVF conference on computer vision and pattern recognition}, 11797--11807.

\bibitem[{Liu et~al.(2021)Liu, Chen, Qin, Dou, and Heng}]{liu2021feddg}
Liu, Q.; Chen, C.; Qin, J.; Dou, Q.; and Heng, P.-A. 2021.
\newblock Feddg: Federated domain generalization on medical image segmentation via episodic learning in continuous frequency space.
\newblock In \emph{Proceedings of the IEEE/CVF conference on computer vision and pattern recognition}, 1013--1023.

\bibitem[{Liu et~al.(2023)Liu, Yin, Qu, and Wang}]{liu2023reducing}
Liu, S.; Yin, S.; Qu, L.; and Wang, M. 2023.
\newblock Reducing domain gap in frequency and spatial domain for cross-modality domain adaptation on medical image segmentation.
\newblock In \emph{Proceedings of the AAAI Conference on Artificial Intelligence}, volume~37, 1719--1727.

\bibitem[{Liu et~al.(2024)Liu, Zhu, Liu, Yu, Chen, and Gao}]{liu2024rolling}
Liu, Y.; Zhu, H.; Liu, M.; Yu, H.; Chen, Z.; and Gao, J. 2024.
\newblock Rolling-Unet: Revitalizing MLP’s Ability to Efficiently Extract Long-Distance Dependencies for Medical Image Segmentation.
\newblock In \emph{Proceedings of the AAAI Conference on Artificial Intelligence}, volume~38, 3819--3827.

\bibitem[{Mukherjee et~al.(2022)Mukherjee, Sarkar, Manich, Labruyere, and Olivo-Marin}]{mukherjee2022domain}
Mukherjee, S.; Sarkar, R.; Manich, M.; Labruyere, E.; and Olivo-Marin, J.-C. 2022.
\newblock Domain adapted multitask learning for segmenting amoeboid cells in microscopy.
\newblock \emph{IEEE Transactions on Medical Imaging}, 42(1): 42--54.

\bibitem[{Niemeijer et~al.(2011)Niemeijer, Xu, Dumitrescu, Gupta, van Ginneken, Folk, and Abr{\`a}moff}]{niemeijer2011automatic}
Niemeijer, M.; Xu, X.; Dumitrescu, A.; Gupta, P.; van Ginneken, B.; Folk, J.; and Abr{\`a}moff, M.~D. 2011.
\newblock Automatic measurement of the arteriolar-to-venular width ratio in digital color fundus photographs.
\newblock \emph{IEEE Transactions on Medical Imaging}, 30(11): 1941--1950.

\bibitem[{Niloy, Bhaumik, and Woo(2024)}]{niloy2024source}
Niloy, F.~F.; Bhaumik, K.~K.; and Woo, S.~S. 2024.
\newblock Source-Free Online Domain Adaptive Semantic Segmentation of Satellite Images Under Image Degradation.
\newblock In \emph{ICASSP 2024-2024 IEEE International Conference on Acoustics, Speech and Signal Processing (ICASSP)}, 7275--7279. IEEE.

\bibitem[{Owen et~al.(2009)Owen, Rudnicka, Mullen, Barman, Monekosso, Whincup, Ng, and Paterson}]{owen2009measuring}
Owen, C.; Rudnicka, A.; Mullen, R.; Barman, S.; Monekosso, D.; Whincup, P.; Ng, J.; and Paterson, C. 2009.
\newblock Measuring retinal vessel tortuosity in 10-year-old children: validation of the computer-assisted image analysis of the retina (CAIAR) program.
\newblock \emph{Investigative ophthalmology \& visual science}, 50(5): 2004--2010.

\bibitem[{Ronneberger, Fischer, and Brox(2015)}]{ronneberger2015u}
Ronneberger, O.; Fischer, P.; and Brox, T. 2015.
\newblock U-net: Convolutional networks for biomedical image segmentation.
\newblock In \emph{Medical image computing and computer-assisted intervention--MICCAI 2015: 18th international conference, Munich, Germany, October 5-9, 2015, proceedings, part III 18}, 234--241. Springer.

\bibitem[{Staal et~al.(2004)Staal, Abramoff, Niemeijer, Viergever, and van Ginneken}]{staal2004ridge}
Staal, J.; Abramoff, M.; Niemeijer, M.; Viergever, M.; and van Ginneken, B. 2004.
\newblock Ridge-based vessel segmentation in color images of the retina.
\newblock \emph{IEEE transactions on medical imaging}, 23(4): 501--509.

\bibitem[{Tishby, Pereira, and Bialek(2000)}]{tishby2000information}
Tishby, N.; Pereira, F.~C.; and Bialek, W. 2000.
\newblock The information bottleneck method.
\newblock \emph{arXiv preprint physics/0004057}.

\bibitem[{Tishby and Zaslavsky(2015)}]{tishby2015deep}
Tishby, N.; and Zaslavsky, N. 2015.
\newblock Deep learning and the information bottleneck principle.
\newblock In \emph{2015 ieee information theory workshop (itw)}, 1--5. IEEE.

\bibitem[{Wang et~al.(2024)Wang, Lin, Wu, Yu, Cheng, and Yan}]{wang2024dtmformer}
Wang, Z.; Lin, X.; Wu, N.; Yu, L.; Cheng, K.-T.; and Yan, Z. 2024.
\newblock DTMFormer: Dynamic Token Merging for Boosting Transformer-Based Medical Image Segmentation.
\newblock In \emph{Proceedings of the AAAI Conference on Artificial Intelligence}, volume~38, 5814--5822.

\bibitem[{Wu et~al.(2023)Wu, Wang, Gu, Lu, Chen, Zhu, Vercauteren, Ourselin, and Zhang}]{wu2023upl}
Wu, J.; Wang, G.; Gu, R.; Lu, T.; Chen, Y.; Zhu, W.; Vercauteren, T.; Ourselin, S.; and Zhang, S. 2023.
\newblock Upl-sfda: Uncertainty-aware pseudo label guided source-free domain adaptation for medical image segmentation.
\newblock \emph{IEEE transactions on medical imaging}.

\bibitem[{Xu et~al.(2021)Xu, Zhang, Zhang, Wang, and Tian}]{xu2021fourier}
Xu, Q.; Zhang, R.; Zhang, Y.; Wang, Y.; and Tian, Q. 2021.
\newblock A fourier-based framework for domain generalization.
\newblock In \emph{Proceedings of the IEEE/CVF conference on computer vision and pattern recognition}, 14383--14392.

\bibitem[{Yang et~al.(2022)Yang, Guo, Chen, and Yuan}]{yang2022source}
Yang, C.; Guo, X.; Chen, Z.; and Yuan, Y. 2022.
\newblock Source free domain adaptation for medical image segmentation with fourier style mining.
\newblock \emph{Medical Image Analysis}, 79: 102457.

\bibitem[{Yang and Soatto(2020)}]{yang2020fda}
Yang, Y.; and Soatto, S. 2020.
\newblock Fda: Fourier domain adaptation for semantic segmentation.
\newblock In \emph{Proceedings of the IEEE/CVF conference on computer vision and pattern recognition}, 4085--4095.

\bibitem[{Ye et~al.(2021)Ye, Zhang, Ouyang, and Yuan}]{ye2021source}
Ye, M.; Zhang, J.; Ouyang, J.; and Yuan, D. 2021.
\newblock Source data-free unsupervised domain adaptation for semantic segmentation.
\newblock In \emph{Proceedings of the 29th ACM international conference on multimedia}, 2233--2242.

\bibitem[{Zhang et~al.(2024)Zhang, Su, Xu, and Jia}]{zhang2024improving}
Zhang, H.; Su, Y.; Xu, X.; and Jia, K. 2024.
\newblock Improving the generalization of segmentation foundation model under distribution shift via weakly supervised adaptation.
\newblock In \emph{Proceedings of the IEEE/CVF Conference on Computer Vision and Pattern Recognition}, 23385--23395.

\bibitem[{Zhou et~al.(2023)Zhou, Ji, Cui, Wang, and Yi}]{zhou2023unsupervised}
Zhou, W.; Ji, J.; Cui, W.; Wang, Y.; and Yi, Y. 2023.
\newblock Unsupervised Domain Adaptation Fundus Image Segmentation via Multi-scale Adaptive Adversarial Learning.
\newblock \emph{IEEE Journal of Biomedical and Health Informatics}.

\end{thebibliography}

\end{document}